\documentclass[journal]{IAENGtran}
\usepackage{wrapfig}
\usepackage{soul}
\usepackage{multirow}
\usepackage{booktabs}
\usepackage{adjustbox}
\usepackage{graphicx}
\usepackage{algorithm}
\usepackage[noend]{algorithmic}
\usepackage{xcolor,colortbl}
\usepackage{color}
\usepackage{textcomp}
\usepackage{amsmath}
\renewcommand{\vec}[1]{\mathbf{#1}}
\usepackage[numbers,sort&compress,square]{natbib}
\definecolor{Gray}{gray}{0.85}
\definecolor{LightCyan}{rgb}{0.88,1,1}

\renewcommand{\vec}[1]{\mathbf{#1}}
\usepackage{epstopdf}
\usepackage{subfloat}
\usepackage{diagbox}
\usepackage[misc]{ifsym}
\usepackage{float}
\usepackage{lipsum}
\usepackage{xfrac}
\usepackage[font=small,labelfont=bf]{caption}
\usepackage[caption = false]{subfig}
\usepackage{caption}
\let\labelindent\relax
\usepackage{enumitem}
\setlist[itemize]{leftmargin=*}
\renewcommand{\arraystretch}{1.2}
\usepackage{lipsum} 
\usepackage{fancyhdr}
\begin{document}\sloppy

\title{
Machine Learning Approach for Skill Evaluation in Robotic-Assisted Surgery
}

\author{Mahtab~J.~Fard, Sattar~Ameri, Ratna~B.~Chinnam, Abhilash~K.~Pandya, Michael~D.~Klein, and~R.~Darin~Ellis
\thanks{Manuscript received July 23, 2016; revised August 8, 2016.}
\thanks{M. J. Fard, S. Ameri R. B. Chinnam, and R.D. Ellis are with the Department
of Industrial and System Engineering, Wayne State University, Detroit, MI, 48202 USA e-mail: {\tt\small {fard@wayne.edu.}}}
\thanks{A. B. Pandya is with Department of Electrical and Computer Engineering, Wayne State University, Detroit, MI, 48202 USA}
\thanks{M. D. Klein is with Department of Surgery, Wayne State University School of Medicine and Pediatric Surgery, Children’s Hospital of Michigan, Detroit, MI, 48202 USA }
}

\maketitle

\begin{abstract}
Evaluating surgeon skill has predominantly been a subjective task. Development of objective methods for surgical skill assessment are of increased interest. Recently, with technological advances such as robotic-assisted minimally invasive surgery (RMIS), new opportunities for objective and automated assessment frameworks have arisen. In this paper, we applied machine learning methods to automatically evaluate performance of the surgeon in RMIS. Six important movement features were used in the evaluation including completion time, path length, depth perception, speed, smoothness and curvature. Different classification methods applied to discriminate expert and novice surgeons. We test our method on real surgical data for suturing task and compare the classification result with the ground truth data (obtained by manual labeling). The experimental results show that the proposed framework can classify surgical skill level with relatively high accuracy of 85.7\%. This study demonstrates the ability of machine learning methods to automatically classify expert and novice surgeons using movement features for different RMIS tasks. Due to the simplicity and generalizability of the introduced classification method, it is easy to implement in existing trainers. .
\end{abstract}
\begin{IAENGkeywords}
Skill assessment, Surgeon skill, Robotic-assisted surgery, Classification, Machine learning.
\end{IAENGkeywords}

\thispagestyle{fancy}
\section{Introduction}
\IAENGPARstart{D}{espite} advances in computer systems and simulation methods, today surgical training is still based on manual assessment involving significant expert monitoring \cite{grantcharov2002assessment, RCS:RCS1766}. For many years, surgical skills have been learned in the operation room under direct supervision of expert surgeons \cite{reznick1993teaching}. These methods are threatened with a lack of consistency, reliability and efficiency due to the subjective nature of experts’ intervention \cite{schout2010validation}. Subjective skill assessment methods are being surrendered for more structured techniques such as Objective Structured Assessment of Technical Skills (OSATS) \cite{martin1997objective}. Using OSATS, an expert surgeon gives scores to surgical trainees based on predefined criteria such as flow of surgery, motion time and final product by observing the surgery in person or watching the recorded video of the operation. 

The new technology innovations such as robotic-assisted minimally invasive surgery (RMIS), open great opportunities for automated objective skill assessment which was not available before  \cite{lalys2014surgical}. The potential of recording motion and video data, has been motivated for a new automatic RMIS skill assessment system \cite{reiley2011review, fard2017eee}. Current systems like da Vinci (Intuitive Surgical, Sunnyvale, CA) \cite{guthart2000intuitivetm} record motion and video data, enabling development of computational models to analyze surgical skills through data-driven approaches \cite{jahanbani2016computational}. However, elaborating such models has always lagged behind. It is, however, quite clear that to develop any framework that automatically evaluates surgical skills, a more rigorous model of surgical procedures is needed \cite{pandya2014review}.  

A number of researchers developed skill assessment methods by decomposing a surgical tasks into pre-defined surgical gestures \cite{mackenzie2001hierarchical}. Most existing work in this area uses statistical approaches such as Hidden Markov Model (HMM) \cite{rosen2002task, tao2012sparse, varadarajan2009data} and descriptive curve coding (DCC) \cite{Ahmidi2015}. Although these methods have the ability to find the underling structure of MIS/RMIS tasks, they are context-based and suffer from requiring large number of training samples and complex parameter tuning, causing in a lack of robustness in the results \cite{tao2012sparse}. On the other hand, most research in objective surgical skill assessment has been focusing entirely on motion features because of their simplicity in implementation and interpretation \cite{Chmarra2010}. Metrics such as operation time, speed, number of hand movements \cite{datta2001use}, force and torque signatures \cite{rosen2001markov}, path length and motion smoothness \cite{Chmarra2010, Cotin2002} have been widely used to identify the relation between the features and surgical tools movement pattern of expert and novices during Laparoscopic surgery \cite{judkins2009objective}.

Although previous work built the foundation of objective surgical skill assessment, the current state of the art has a few shortcomings. First, they mostly focus on descriptive statistical methods to show the dependency of surgical skill level and GFMs. However, these measures alone are not an adequate proficiency measurement. More advanced techniques such as data mining and machine learning algorithms need to be applied \cite{Dreiseitl2005}. While machine learning techniques have been used extensively in other fields \cite{ameri2016survival, fard2016early, mahtabIEOM2015} because of their advantages over traditional statistical methods such as robustness, better prediction ability and higher tolerance violence of assumptions (e. c. normality or undependability of data) \cite{Murphy2012, 7564399}, but it is only recently that these methods have been considered to analyze RMIS tasks  \cite{Kassahun2015, fard2016skill}. Additionally, human factor study should be developed to have better understanding of this aspect in surgical training \cite{ellis2012management, yang2012using}. Thus, developing quantitative classification techniques that can automatically and accurately evaluate surgical skills needs to be investigated.

\section{Surgical Skill Evaluation Framework}
In this paper, we develop a predictive framework for objective skill assessment based on the trajectory movement of the surgical robot arms. For this, we quantify surgical task by extracting movement features from the raw motion data for suturing task. Different classifiers, including logistic regression and support vector machines have been applied. The classifier with the high accuracy can be used to automatically predict the skill level of surgeon. 

\begin{figure*}[htbp]
\centering
\scalebox{.8}{\includegraphics{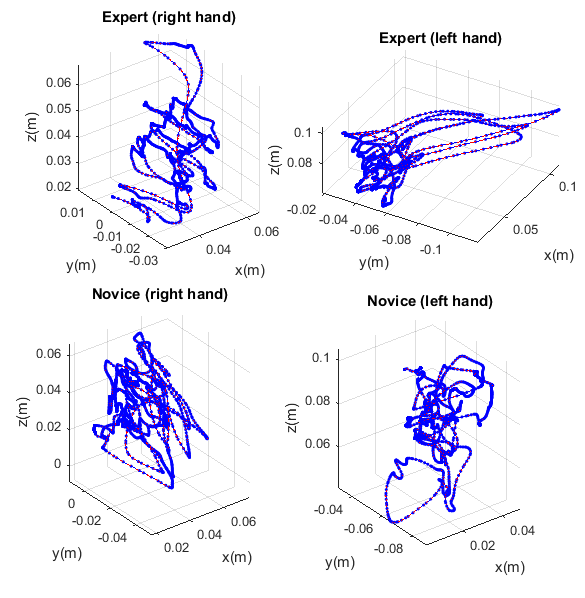}}
\caption{Illustration of the Cartesian position plots for an expert and a novice surgeon doing suturing on the da Vinci surgical robot..}
\label{one}
\end{figure*}

\subsection{Features Extraction}
Surgical tasks have different characteristics, such as smoothness, straightness or response orientation, which account for competence while relying only on instrument motion. For instance, studies have shown that the tool motion of an experienced surgeon has more clearly defined features than that of a less experienced surgeon while performing the same task \cite{lin2006towards}. Figures \ref{one} illustrates the Cartesian position plots of an expert and a novice surgeon doing four throw suturing on the da Vinci surgical robot.

In order to transform these parameters into quantitative metrics, we applied kinematic analysis theory that has been successfully used in previous works to study psychomotor skills \cite{Chmarra2010}. Metrics such as task completion time, length of path, depth perception and velocity can show some aspect of surgeon's dexterity. However, other aspects such as smoothness, curvature, torsion and complexity of the motion need to be quantified.  In the following, we explain the six important features from the clinical point of view.
\vspace{0.5em}
\begin{itemize}
\item {\textbf{\em Time to Complete (TTC)}}: is defined as total time required to complete the task, measured in seconds.
\vspace{0.5em}
\item {\textbf{\em Path Length (PL)}}: is the length of the curve described by the tip of the instrument while performing the task (in $cm$). We calculate it using sum of all consecutive pairs' Euclidean distance.
\vspace{0.5em}
\item {\textbf{\em Depth Perception (DP)}}: is the total distance traveled by the instrument along its axis (in cm). 
\vspace{0.5em}
\item {\textbf{\em Speed}}: can be defined as the magnitude of velocity and calculated as the rate of position change from previous time step as $\sfrac{dis(p_i,p_(i-1))}{(\Delta t_i)}$, where $dis(p_i,p_(i-1))$ can be calculated as a Euclidean distance between $i^{th}$ point and of ${(i-1)}^{th}$ point (in $\sfrac{cm}{s}$). Given that the time difference between two consecutive frames in our signal is constant, $\Delta t_i$  is equal to 1.
\vspace{0.5em}
\item {\textbf{\em Motion Smoothness}}: is a measure of the rhythmic pattern of acceleration and deceleration. Smoothness has most often been based on minimizing jerk, the third time derivative of position, which represents a change in acceleration (in $\sfrac{cm}{s^3}$). 
\vspace{0.5em}
\item {\textbf{\em Curvature}}: measures the straightness of the path and is calculated at each point by the following equation \cite{Chmarra2010}
$$\kappa_i=\frac{v_i\times a_i}{v_i^{3}}$$
where $v_i$ and $a_i$  are instantaneous velocity and acceleration of the instrument tips respectively, which can be calculated directly by computing the first and second derivatives of the positions of the instrument tips. The curvature measures how fast a curve is changing direction at a given point. For straight and smooth movement, the mean of curvature is close to zero, while larger values indicate curved an jerky movements.
\end{itemize}

\subsection{Surgical Skill Classification}
Features that are extracted in the previous section are used to quantify the movement pattern of surgeons with different levels of dexterity. Our aim is to build a discriminative model to differentiate between surgeons with different levels of expertise while doing RMIS tasks. Surgeons are categorized into two skill levels, expert and novice. Thus, this is a binary classification problem that can be resolved by machine learning algorithms. In particular, we compared two frequently used machine learning techniques, Logistic regression \cite{Kleinbaum2010} and Support Vector Machine \cite{Saitta1995}. 

\vspace{0.4em}
\subsubsection{Logistic Regression (LR)}
One of the well-established statistical models is the Logistic regression where the dependent variable is categorical. In this model, logit transformation of a linear combination of features is used to resolve a binary classification problem. Formally, the logistic regression model can be formalized as 
\begin{equation}
p(x)=\frac{1}{1+e^{(-(\beta_0+\beta.x))}}
\label{LR}
\end{equation}
where $\beta$ is the coefficient for corresponding $x$ feature and $p(x)$ is the probability of belonging to one of the classes.

\vspace{0.4em}
\subsubsection{Support Vector Machine (SVM)}
Support vector machine (SVM) is one of the important classification method that constructs a hyperplane and tries to maximize the margin that separate two class of data shows as $\sfrac{2}{||\vec{\omega}||}$. The ability to learn the non-linear separable function by mapping the data to a higher dimensional space makes this classifier unbeatable for some classification problems. Linear SVM can be formalized as

\begin{equation}
\begin{aligned}
& {\text{minimize}}
& & \frac{2}{||\vec{\omega}||}\\
& \text{subject to}
& & y_i(\omega\cdot x_i+b)\geq 1 & \forall i=1, ..., n
\end{aligned}
\label{svm}
\end{equation}
where $y_i$ is the class label for $i^{th}$ data. In order to solve the non-linear classification problem, SVM uses a kernel transformation. In this study we applied radial basis function (RBF) which is one of the most popular kernel functions used in SVM \cite{Vapnik1998}, defined as

\begin{equation}
K(x_i, x_j)=e^{(-\gamma{||x_i-x_j||}^2)}
\label{RBF}
\end{equation}
where $\gamma$ controls the width of RBF function.

\section{Experimental Results}
In this section, we describe the experimental method including the dataset and feature extraction in detail for each surgical task. We also explain the performance metrics that we used to evaluate the proposed automated surgical skill assessment framework.

\subsection{Dataset}
\begin{figure}[!h]
\centering
\scalebox{.6}{\includegraphics{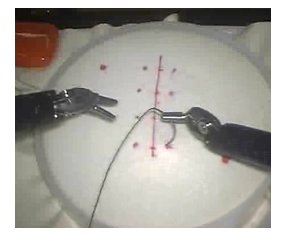}}
\caption{Snapshot of the suturing task in robotic-assisted minimally invasive surgery.}
\label{fig:2task}
\end{figure}

We implemented our framework on real robotic surgery data presented in \cite{gaojhu}. This is comprised of data for suturing task (see Figure \ref{fig:2task}). Eight right-handed surgeons with different skill levels performed suturing around 5 times. We analyze kinematic data captured using the API of the da Vinci robot at 30 Hz to extract features. The data includes manual annotation for surgeon’s skill based on a global rating score (GRS). Surgeons are divided into two categories of experts and novices based on their scores. 
\vspace{0.5em}

Feature are extracted for both hands using Cartesian positions of right and left patient-side manipulator end-effectors of da Vinci arms. Before computing the features, the raw data were filtered using a local regression weighted linear least square method, which reduces the noise in signal data and keeps the detail of the pattern. Speed, motion smoothness and curvature are temporal features and were calculated for each point in data. Therefore, the descriptive statistics including mean and standard deviation are derived for these features. Finally, a total of 17 features are derived from each trajectory. Then, we employed principle component analysis (PCA), a dimensionality reduction technique that is based on an orthogonal transformation to reduce the set of possible correlated features to a smaller set of uncorrelated features that are linear combinations of the original features \cite{Abdi2010}.

\subsection{Performance Evaluation}
Classifier validation was conducted using two model validation schemas as suggested in \cite{gaojhu}. The first is leave-one-super-trial-out (LOSO), where one trial for each one of the surgeons is left out for testing. The second is leave-one-user-out (LOUO), where we leave out all the trials from one surgeon for testing. While the first validation method evaluates the robustness of a method for repeating a task by leaving out one trial for all subjects, the second setup evaluates the robustness of a method when a subject (i.e., surgeon) is not previously seen in the training data. The performance of the different classification methods was determined by classification accuracy, which is expressed in terms of percentage of subjects in the test set that are classified correctly.

The results of performing two classification methods, logistic regression and SVM using LOSO and LOUO for suturing, are shown in Table \ref{one}. The best accuracy was obtained for the combination of all movement features for both tasks. Table \ref{one} shows that the best overall accuracy that has been achieved for suturing is 85.7\% in LOSO and 71.9\% in LOUO. Results also show that logistic regression for LOSO and SVM for LOUO model validation schema provide the best classification performance. 
\vspace{0.3em}
\begin{table}[!htbp]
\centering
\caption{Classification accuracy for skill level evaluation in suturing task using Logistic regression (LR) and SVM for two validation schema LOSO and LOUO (best accuracy is highlighted in bold).}
\renewcommand{\arraystretch}{1.2}
\begin{tabular}{cccc}
 &  & \textbf{LOSO} & \textbf{LOUO}\tabularnewline
\hline 
\hline 
\multirow{2}{*}{\textbf{Novices}} & LR & \textbf{79.3\%} & 66.7\%\tabularnewline
 & SVM & 47.1\% & \textbf{67.9\%}\tabularnewline
\hline 
\multirow{2}{*}{\textbf{Experts}} & LR & \textbf{83.5\%} & 69.8\%\tabularnewline
 & SVM & 61.3\% & \textbf{71.2\%}\tabularnewline
\hline 
\multirow{2}{*}{\textbf{Overall}} & LR & \textbf{85.7\%} & 70.5\%\tabularnewline
 & SVM & 58.7\% & \textbf{72.1\%}\tabularnewline
\hline 
\end{tabular}
\label{one}
\end{table}

From the results shown in Table \ref{one} , the classification accuracy improves when combination of spatial and curvature features are used. This is consistent with previous studies which emphasized that task completion time and distance traveled are insufficient to explain all aspects of surgical assessment. The features that have been used in this study can be applied globally on RMIS tasks as they have the potential to identify additional aspects of surgeon skill level which cannot be quantified by task completion time and distance traveled alone.

Table \ref{one} also shows that for almost all experiments, the best overall accuracy obtained was from logistic regression for LOSO schema while SVM gives the best result for LOUO. It should be noted that, the result of LOUO provides an insight into the ability of the algorithms to evaluate the skill level of a surgeon that was unseen during the training phase. Therefore, we can conclude that the underlying pattern of different surgeons with same skill level is not linear. Therefore, SVM with a nonlinear kernel, such as the one we applied in this study (RBF), has better classification ability to assess the skill level of surgeons who are not previously seen in the training data. In other words, SVM has more generalizability in this context. Interestingly, experts can be classified with higher accuracy compared to novices due to the consistency in the values of movement features for experts. It is also important to mention that the overall accuracy of surgical skill classification decreases 16\% when we switch from LOSO validation schema to LOUO. This suggests that surgeons with the same level of expertise perform suturing in a more similar way. 

\section{Conclusion}
This study demonstrates the ability of machine learning methods to automatically distinguish between expert and novice performance in robotic-assisted surgical tasks. It is generally accepted that not only the skill level of surgeon vary, but also each surgical task has different levels of complexity. This complexity is not only captured through the features extracted from trajectory movement data, but also through more advanced machine learning methods that are needed to model the underlying pattern of surgical skill level. The results presented in this paper could form a basis for decision support tools that effectively, objectively and automatically evaluate surgeon’s dexterity and provide more personalized skill assessment and online feedback to trainees based on their performance. Furthermore, the proposed method can be applied on a more granular level of tasks in robotic-assisted surgery, such as surgical gestures, to provide more insight into the skill level differences of surgeons. Future research could focus on performing more validation studies with a larger number of participants for different surgical tasks, which would yield a larger training set with the potential for improving the classification result. 

\bibliographystyle{BibTeXtran}  
\small \bibliography{IntJCARS}  

\begin{thebibliography}{10}
\providecommand{\url}[1]{#1}
\csname url@samestyle\endcsname
\providecommand{\newblock}{\relax}
\providecommand{\bibinfo}[2]{#2}
\providecommand{\BIBentrySTDinterwordspacing}{\spaceskip=0pt\relax}
\providecommand{\BIBentryALTinterwordstretchfactor}{4}
\providecommand{\BIBentryALTinterwordspacing}{\spaceskip=\fontdimen2\font plus
\BIBentryALTinterwordstretchfactor\fontdimen3\font minus
  \fontdimen4\font\relax}
\providecommand{\BIBforeignlanguage}[2]{{%
\expandafter\ifx\csname l@#1\endcsname\relax
\typeout{** WARNING: IEEEtran.bst: No hyphenation pattern has been}%
\typeout{** loaded for the language `#1'. Using the pattern for}%
\typeout{** the default language instead.}%
\else
\language=\csname l@#1\endcsname
\fi
#2}}
\providecommand{\BIBdecl}{\relax}
\BIBdecl

\bibitem{grantcharov2002assessment}
T.~P. Grantcharov, L.~Bardram, P.~Funch-Jensen, and J.~Rosenberg, ``Assessment
  of technical surgical skills,'' \emph{The European journal of surgery}, vol.
  168, no.~3, pp. 139--144, 2002.

\bibitem{RCS:RCS1766}
\BIBentryALTinterwordspacing
M.~J. Fard, A.~K. Pandya, R.~B. Chinnam, M.~D. Klein, and R.~D. Ellis,
  ``Distance-based time series classification approach for task recognition
  with application in surgical robot autonomy,'' \emph{The International
  Journal of Medical Robotics and Computer Assisted Surgery}, pp. n/a--n/a,
  2016, rCS-16-0026.R2. [Online]. Available:
  \url{http://dx.doi.org/10.1002/rcs.1766}
\BIBentrySTDinterwordspacing

\bibitem{reznick1993teaching}
R.~K. Reznick, ``Teaching and testing technical skills,'' \emph{The American
  journal of surgery}, vol. 165, no.~3, pp. 358--361, 1993.

\bibitem{schout2010validation}
B.~Schout, A.~Hendrikx, F.~Scheele, B.~Bemelmans, and A.~Scherpbier,
  ``Validation and implementation of surgical simulators: a critical review of
  present, past, and future,'' \emph{Surgical endoscopy}, vol.~24, no.~3, pp.
  536--546, 2010.

\bibitem{martin1997objective}
J.~Martin, G.~Regehr, R.~Reznick, H.~MacRae, J.~Murnaghan, C.~Hutchison, and
  M.~Brown, ``Objective structured assessment of technical skill (osats) for
  surgical residents,'' \emph{British Journal of Surgery}, vol.~84, no.~2, pp.
  273--278, 1997.

\bibitem{lalys2014surgical}
F.~Lalys and P.~Jannin, ``Surgical process modelling: a review,''
  \emph{International journal of computer assisted radiology and surgery},
  vol.~9, no.~3, pp. 495--511, 2014.

\bibitem{reiley2011review}
C.~E. Reiley, H.~C. Lin, D.~D. Yuh, and G.~D. Hager, ``Review of methods for
  objective surgical skill evaluation,'' \emph{Surgical endoscopy}, vol.~25,
  no.~2, pp. 356--366, 2011.

\bibitem{fard2017eee}
M.~J. Fard, S.~Ameri, R.~B. Chinnam, and R.~D. Ellis, ``Soft boundary approach
  for unsupervised gesture segmentation in robotic-assisted surgery,''
  \emph{IEEE Robotics and Automation Letters}, vol.~2, no.~1, pp. 171--178, Jan
  2017.

\bibitem{guthart2000intuitivetm}
G.~Guthart and J.~K. Salisbury~Jr, ``The intuitivetm telesurgery system:
  Overview and application.'' in \emph{ICRA}, 2000, pp. 618--621.

\bibitem{jahanbani2016computational}
M.~Jahanbani~Fard, ``Computational modeling approaches for task analysis in
  robotic-assisted surgery,'' 2016.

\bibitem{pandya2014review}
A.~Pandya, L.~A. Reisner, B.~King, N.~Lucas, A.~Composto, M.~Klein, and R.~D.
  Ellis, ``A review of camera viewpoint automation in robotic and laparoscopic
  surgery,'' \emph{Robotics}, vol.~3, no.~3, pp. 310--329, 2014.

\bibitem{mackenzie2001hierarchical}
L.~MacKenzie, J.~Ibbotson, C.~Cao, and A.~Lomax, ``Hierarchical decomposition
  of laparoscopic surgery: a human factors approach to investigating the
  operating room environment,'' \emph{Minimally Invasive Therapy \& Allied
  Technologies}, vol.~10, no.~3, pp. 121--127, 2001.

\bibitem{rosen2002task}
J.~Rosen, M.~Solazzo, B.~Hannaford, and M.~Sinanan, ``Task decomposition of
  laparoscopic surgery for objective evaluation of surgical residents' learning
  curve using hidden markov model,'' \emph{Computer Aided Surgery}, vol.~7,
  no.~1, pp. 49--61, 2002.

\bibitem{tao2012sparse}
L.~Tao, E.~Elhamifar, S.~Khudanpur, G.~D. Hager, and R.~Vidal, ``Sparse hidden
  markov models for surgical gesture classification and skill evaluation,'' in
  \emph{Information Processing in Computer-Assisted Interventions}.\hskip 1em
  plus 0.5em minus 0.4em\relax Springer, 2012, pp. 167--177.

\bibitem{varadarajan2009data}
B.~Varadarajan, C.~Reiley, H.~Lin, S.~Khudanpur, and G.~Hager, ``Data-derived
  models for segmentation with application to surgical assessment and
  training,'' in \emph{Medical Image Computing and Computer-Assisted
  Intervention--MICCAI 2009}.\hskip 1em plus 0.5em minus 0.4em\relax Springer,
  2009, pp. 426--434.

\bibitem{Ahmidi2015}
N.~Ahmidi, P.~Poddar, J.~D. Jones, S.~S. Vedula, L.~Ishii, G.~D. Hager, and
  M.~Ishii, ``{Automated objective surgical skill assessment in the operating
  room from unstructured tool motion in septoplasty},'' \emph{International
  Journal of Computer Assisted Radiology and Surgery}, vol.~10, no.~6, pp.
  981--991, 2015.

\bibitem{Chmarra2010}
M.~K. Chmarra, S.~Klein, J.~C.~F. {De Winter}, F.~W. Jansen, and J.~Dankelman,
  ``{Objective classification of residents based on their psychomotor
  laparoscopic skills},'' \emph{Surgical Endoscopy and Other Interventional
  Techniques}, vol.~24, no.~5, pp. 1031--1039, 2010.

\bibitem{datta2001use}
V.~Datta, S.~Mackay, M.~Mandalia, and A.~Darzi, ``The use of electromagnetic
  motion tracking analysis to objectively measure open surgical skill in the
  laboratory-based model,'' \emph{Journal of the American College of Surgeons},
  vol. 193, no.~5, pp. 479--485, 2001.

\bibitem{rosen2001markov}
J.~Rosen, B.~Hannaford, C.~G. Richards, and M.~N. Sinanan, ``Markov modeling of
  minimally invasive surgery based on tool/tissue interaction and force/torque
  signatures for evaluating surgical skills,'' \emph{Biomedical Engineering,
  IEEE Transactions on}, vol.~48, no.~5, pp. 579--591, 2001.

\bibitem{Cotin2002}
S.~Cotin, N.~Stylopoulos, M.~P. Ottensmeyer, P.~F. Neumann, D.~W. Rattner, and
  S.~Dawson, ``{Metrics for Laparoscopic Skills Trainers: The Weakest Link!}''
  in \emph{Medical Image Computing and Computer-Assisted Intervention--MICCAI
  2002}, vol. 2488, 2002, pp. 35--43.

\bibitem{judkins2009objective}
T.~N. Judkins, D.~Oleynikov, and N.~Stergiou, ``Objective evaluation of expert
  and novice performance during robotic surgical training tasks,''
  \emph{Surgical endoscopy}, vol.~23, no.~3, pp. 590--597, 2009.

\bibitem{Dreiseitl2005}
S.~Dreiseitl and M.~Binder, ``{Do physicians value decision support? A look at
  the effect of decision support systems on physician opinion},''
  \emph{Artificial Intelligence in Medicine}, vol.~33, no.~1, pp. 25--30, 2005.

\bibitem{ameri2016survival}
S.~Ameri, M.~J. Fard, R.~B. Chinnam, and C.~K. Reddy, ``Survival analysis based
  framework for early prediction of student dropouts,'' in \emph{Proceedings of
  the 25th ACM International on Conference on Information and Knowledge
  Management}.\hskip 1em plus 0.5em minus 0.4em\relax ACM, 2016, pp. 903--912.

\bibitem{fard2016early}
M.~J. Fard, S.~Chawla, and C.~K. Reddy, ``Early-stage event prediction for
  longitudinal data,'' in \emph{Pacific-Asia Conference on Knowledge Discovery
  and Data Mining}.\hskip 1em plus 0.5em minus 0.4em\relax Springer, 2016, pp.
  139--151.

\bibitem{mahtabIEOM2015}
M.~J. Fard, S.~Ameri, and A.~Zeinal~Hamadani, ``Bayesian approach for early
  stage reliability prediction of evolutionary products,'' in \emph{Proceedings
  of the International Conference on Operations Excellence and Service
  Engineering}.\hskip 1em plus 0.5em minus 0.4em\relax Orlando, Florida, USA,
  2015, pp. 361--371.

\bibitem{Murphy2012}
K.~P. Murphy, \emph{{Machine Learning: A Probabilistic Perspective}}.\hskip 1em
  plus 0.5em minus 0.4em\relax The MIT Press, aug 2012.

\bibitem{7564399}
M.~J. Fard, P.~Wang, S.~Chawla, and C.~K. Reddy, ``A bayesian perspective on
  early stage event prediction in longitudinal data,'' \emph{IEEE Transactions
  on Knowledge and Data Engineering}, vol.~28, no.~12, pp. 3126--3139, Dec
  2016.

\bibitem{Kassahun2015}
Y.~Kassahun, B.~Yu, A.~T. Tibebu, D.~Stoyanov, S.~Giannarou, J.~H. Metzen, and
  E.~{Vander Poorten}, ``{Surgical robotics beyond enhanced dexterity
  instrumentation: a survey of machine learning techniques and their role in
  intelligent and autonomous surgical actions},'' \emph{International Journal
  of Computer Assisted Radiology and Surgery}, vol.~11, no.~4, pp. 553--568,
  2016.

\bibitem{fard2016skill}
M.~J. Fard, S.~Ameri, and R.~D. Ellis, ``Toward personalized training and skill
  assessment in robotic minimally invasive surgery,'' \emph{arXiv preprint
  arXiv:1610.07245v2}, 2016.

\bibitem{ellis2012management}
R.~D. Ellis, M.~J. Fard, K.~Yang, W.~Jordan, N.~Lightner, and S.~Yee,
  ``Management of medical equipment reprocessing procedures: A human
  factors/system reliability perspective,'' in \emph{Advances in Human Aspects
  of Healthcare}.\hskip 1em plus 0.5em minus 0.4em\relax CRC Press, 2012, pp.
  689--698.

\bibitem{yang2012using}
K.~Yang, N.~Lightner, S.~Yee, M.~J. Fard, and W.~Jordan, ``Using computerized
  technician competency validation to improve reusable medical equipment
  reprocessing system reliability,'' in \emph{Advances in Human Aspects of
  Healthcare}.\hskip 1em plus 0.5em minus 0.4em\relax CRC Press, 2012, pp.
  556--564.

\bibitem{lin2006towards}
H.~C. Lin, I.~Shafran, D.~Yuh, and G.~D. Hager, ``Towards automatic skill
  evaluation: Detection and segmentation of robot-assisted surgical motions,''
  \emph{Computer Aided Surgery}, vol.~11, no.~5, pp. 220--230, 2006.

\bibitem{Kleinbaum2010}
D.~G. Kleinbaum and M.~Klein, \emph{{Logistic Regression}}, ser. Statistics for
  Biology and Health.\hskip 1em plus 0.5em minus 0.4em\relax New York, NY:
  Springer New York, 2010.

\bibitem{Saitta1995}
L.~Saitta, ``{Suppor Vector Networks},'' \emph{Machine Learning}, vol.~20, pp.
  273--297, 1995.

\bibitem{Vapnik1998}
V.~Vapnik, \emph{{Statistical Learning Theory}}, 1998.

\bibitem{gaojhu}
Y.~Gao, S.~S. Vedula, C.~E. Reiley, N.~Ahmidi, B.~Varadarajan, H.~C. Lin,
  L.~Tao, L.~Zappella, B.~B{\'e}jar, D.~D. Yuh \emph{et~al.}, ``{JHU-ISI}
  gesture and skill assessment working set ({JIGSAWS}): A surgical activity
  dataset for human motion modeling,'' in \emph{Modeling and Monitoring of
  Computer Assisted Interventions ({M2CAI})– {MICCAI} Workshop}, 2014.

\bibitem{Abdi2010}
H.~Abdi and L.~J. Williams, ``{Principal component analysis},'' \emph{Wiley
  Interdisciplinary Reviews: Computational Statistics}, vol.~2, no.~4, pp.
  433--459, jul 2010.

\end{thebibliography}

\end{document}